\documentclass{article}

\usepackage{arxiv}
\usepackage{amsmath}
\usepackage[utf8]{inputenc} 
\usepackage[T1]{fontenc}    
\usepackage[colorlinks=true,linkcolor=blue,citecolor=green, urlcolor=blue]{hyperref}       
\usepackage{url}            
\usepackage{booktabs}       
\usepackage{amsfonts}       
\usepackage{nicefrac}       
\usepackage{microtype}      
\usepackage{authblk} 
\usepackage{lipsum}
\usepackage{tabularx}
\usepackage{graphicx}
\usepackage{adjustbox}
\usepackage{float}
\usepackage[numbers,sort&compress]{natbib}
\graphicspath{ {./images/} }
\usepackage{cleveref}
\crefname{figure}{}{}
\crefname{section}{Sec.}{Secs.}
\Crefname{section}{Section}{Sections}
\Crefname{table}{Table}{Tables}
\crefname{table}{Tab.}{Tabs.}
\renewcommand{\thefigure}{\arabic{figure}}
\renewcommand{\thetable}{\arabic{table}}

\DeclareFontFamily{U}{matha}{\hyphenchar\font45}
\DeclareFontShape{U}{matha}{m}{n}{
      <5> <6> <7> <8> <9> <10> gen * matha
      <10.95> matha10 <12> <14.4> <17.28> <20.74> <24.88> matha12
      }{}
\DeclareSymbolFont{matha}{U}{matha}{m}{n}

\DeclareMathSymbol{\Lt}{3}{matha}{"CE}
\DeclareMathSymbol{\Gt}{3}{matha}{"CF}

\title{
Transformers for Molecular Property Prediction: Domain Adaptation efficiently improves performance.
}

\author[1$\dagger$]{Afnan Sultan}
\author[2$\dagger$]{Max Rausch-Dupont}
\author[2,3]{Shahrukh Khan}
\author[3]{Olga Kalinina}
\author[2$\S$]{Dietrich Klakow}
\author[1, 3$\S$]{Andrea Volkamer}

\affil[1]{Data Driven Drug Design, Center for Bioinformatics, Saarland University, Saarbrücken, Germany}
\affil[2]{Spoken Language Systems, Saarland Informatics Campus, Saarland University, Saarbrücken, Germany}
\affil[3]{Medical Faculty, Saarland University, 66421, Homburg, Saarland, Germany; Center for Bioinformatics, Saarland University, 66123,
Saarbrücken, Saarland, Germany}

\renewcommand{\today}{}

\begin{document}
\maketitle

\renewcommand{\thefootnote}{} 
\footnotetext{$\dagger$ Co-first authors.}
\footnotetext{$\S$ Co-corresponding authors.}

\begin{abstract}
Over the past six years, molecular transformer models have become an integral part of the computational toolbox for drug discovery. Most existing models are pre-trained on millions to billions of molecules from large-scale unlabeled datasets such as ZINC or ChEMBL. However, the extent to which such large-scale pre-training improves molecular property prediction remains unclear.

This study investigates the potential of transformer models for molecular property prediction while addressing their current limitations. We explore strategies to enhance performance, including the influence of pre-training dataset size and the benefits of domain adaptation through chemically informed objectives.
Our results show that increasing the pre-training dataset beyond approximately 400K–800K molecules does not improve performance across seven datasets covering five ADME endpoints: lipophilicity, permeability, solubility (two datasets), microsomal stability (two datasets), and plasma protein binding. In contrast, applying domain adaptation on a small number of domain-relevant molecules ($\leq 4K$) using multi-task regression of physicochemical properties significantly improves model performance across all datasets (P-value < 0.001).

Furthermore, we find that a model pre-trained on $\sim$400K molecules and adapted on a small domain-specific dataset outperforms larger-scale transformer models like MolFormer and performs comparably to MolBERT. Benchmarking these models alongside Random Forest (RF) baselines using physicochemical descriptors and Morgan fingerprints reveals that incorporating chemically and physically informed features consistently leads to superior performance, regardless of whether used with traditional or transformer-based architectures.

While traditional models, such as RF, remain strong baselines, this study identifies concrete practices that significantly enhance the performance of transformer models. In particular, aligning pre-training and adaptation with chemically meaningful tasks and domain-relevant data offers a promising path forward for future advancements in molecular property prediction. 

Our models are available on HuggingFace to allow for easy use and adaptation at 
\url{https://huggingface.co/collections/UdS-LSV/domain-adaptation-molecular-transformers-6821e7189ada6b7d0a5b62d4}.

\end{abstract}

\newpage

\tableofcontents

\newpage

\section{Introduction}

\textbf{Molecular property prediction (MPP)} is a task at the heart of diverse cheminformatics and drug design challenges. Molecular properties can range from inherent physical features of the molecule like lipophilicity or solubility to more complex results of the physiological properties like toxic effects of the molecule on an organism \cite{wu2018moleculenet}. Supervised learning (SL) has been used to map predefined or heuristic molecular descriptors to such properties \cite{wu2018moleculenet, deng2023systematic}. However, freely available datasets for MPP tasks usually consist of only a few hundred to a few thousand molecules \cite{wu2018moleculenet, fang2023prospective} due to the complex and expensive experimental processes to generate the data \cite{dimasi2010trends}. 
Existing property prediction methods suffer from limitations regarding data representation. On the one end, human-made descriptors like predefined fingerprints require expert knowledge and are restricted to known rules or patterns \cite{wigh2022review}. On the other end, data-driven methods like deep learning require a large amount of labeled data \cite{rong2020self}.

Self-supervised learning (SSL) has been used as an alternative to learning from labeled data as in supervised learning \cite{de1993learning}. In SSL, the model is initially trained on large unlabeled data to learn intrinsic relationships within the input. These relationships can be obtained by using tasks like recovering parts of the input using information from other parts of the input \cite{liu2021self, gui2024survey}. Such an SSL model is then thought of as a foundation model that can be generalized to different downstream tasks \cite{gui2024survey, xie2022self}. The past decade has seen a breakthrough in the field of Natural Language Processing (NLP) with the introduction of the SSL-based transformer model \cite{vaswani2017attention}, which has inspired multiple works to adopt similar schemes for sequence-based representations of molecules \cite{grisoni2023chemical, sultan2024transformers, liao2024words}. 

The \textbf{transformer model} \cite{vaswani2017attention} is a sequence-to-sequence model composed of an encoder-decoder architecture and trained on the next token prediction (NTP) objective. In this objective, the model is optimized such that the decoder model correctly predicts the next token (i.e., subword) given the previous tokens. While the transformer model was built for machine translation, BERT (Bidirectional Encoder Representations from Transformers) \cite{devlin-etal-2019-bert} introduced the concept of transfer learning. BERT was pre-trained on a large corpus of generic, unlabeled data (e.g., Wikipedia), and then fine-tuned on smaller, labeled downstream datasets to generalize across various tasks. One of the pre-training objectives used in BERT \cite{devlin-etal-2019-bert} is masked language modeling (MLM). In this objective, a percentage of the tokens of each sequence is masked randomly, and the model is optimized to correctly predict these masked tokens. 

Although the transfer learning scheme employed in BERT has yielded promising results for numerous tasks \cite{devlin-etal-2019-bert, brown2019guacamol}, its effectiveness is limited when applied to downstream tasks that fall outside the domain of the pre-training corpus. 
For example, a model trained on Wikipedia will not be able to capture the nuances of medical or legal languages. 
To address this, specialized models have been pre-trained on domain-specific corpus rather than general-purpose \cite{rasmy2021med, chalkidis-etal-2020-legal}. 
However, such specialized models require a large and diverse corpus from the specific domain to accurately capture its nuances.
When a sufficient domain-specific corpus is not available, an intermediate step, known as \textbf{domain adaptation (DA)}, is often performed. In this process, the generic model is further trained on the available unlabeled domain-specific corpus \cite{chalkidis-etal-2020-legal}.
The further training step in DA is expected to update the model's weights, integrating knowledge from the desired domain alongside its already established knowledge from the pre-training data. 

\textbf{Molecular transformer models} used for property prediction tasks have predominantly followed the pre-training $\rightarrow$ fine-tuning scheme, with few exceptions. For instance, K-BERT \cite{wu2022knowledge} was initially pre-trained on a dataset that lacked chiral information. To address this limitation, the model was later further-trained on a chirality dataset with an additional chirality classification objective.
Many models have employed generalizable domain-specific objectives during pre-training. For example, MolBERT \cite{fabian2020molecular} and ChemBERTa-2 \cite{ahmad2022chemberta2} trained their models to predict around 210 physicochemical properties for each molecule in the pre-training dataset (referred to as PhysChemPred and MTR in the two manuscripts, respectively). K-BERT \cite{wu2022knowledge} also trained their model to predict features per atom and a structural vector of the molecule calculated using the MACCS structural keys algorithm. Another widely used domain-specific objective is contrastive learning (CL) \cite{chen2020simple}, which has been adopted by models such as Chemformer \cite{irwin2022chemformer}, MolBERT \cite{fabian2020molecular}, and Transformer-CNN \cite{karpov2020transformer} using SMILES sequence augmentation.
While domain-relevant objectives are argued to improve performance \cite{fabian2020molecular, ahmad2022chemberta2}, interpretability \cite{wu2022knowledge}, and model stability \cite{maziarka2020molecule}, they can be computationally expensive to implement on large pre-training datasets \cite{ahmad2022chemberta2}. To date, models incorporating domain-specific objectives during the computationally demanding pre-training step.

Although current molecular transformer models have been pre-trained on millions to billions of molecules, investigations have shown that increasing the pre-training dataset size does not consistently lead to improved predictions of molecular properties \cite{sultan2024transformers}. 
Several studies have explored the impact of increasing pre-training dataset size \cite{ross2022large, ahmad2022chemberta2, chithrananda2020chemberta, chen2017sampling}, but this has not yet been done exhaustively, and these studies often lack distribution or significance analysis, which is crucial for ensuring the robustness.
For example, pre-training on molecules combined from different databases, like ZINC and PubChem as in MolFormer \cite{ross2022large} or ZINC, ChEMBL, and PubChem as in Chen \textit{et. al.}. \cite{chen2017sampling}, did not show noticeable differences in performance compared to using a single database. 
Additionally, it has recently been demonstrated in the field of material science that large databases often contain a substantial amount of redundant information \cite{li2023exploiting}, while for the protein language modeling field, a small but diverse pre-training dataset was sufficient to improve performance on downstream analysis \cite{marquet2024expert}. 

These observations \cite{sultan2024transformers, li2023exploiting, marquet2024expert} let us hypothesize that a current limitation in witnessing the power of transformer models in the molecular property area might be the redundancy in the pre-training dataset and its limitedness in capturing nuances and patterns that are causative for the measured target value of the downstream molecules. 
In this work, we therefore try to answer the following research questions:
\begin{enumerate}
    \item How does increasing pre-training dataset size affect molecular property prediction?
    \item How does domain adaptation with different objectives affect molecular property prediction?
    \item What is the most efficient training approach without compromising performance?
    \item How does domain adaptation compare to currently existing transformer and baseline models?
\end{enumerate}

We hypothesize that performing domain adaptation (DA), as represented by the further-training step, will introduce the model to more relevant datapoints, thereby improving its performance. To perform DA, the domain-specific molecules are selected as the new unlabeled data for further training. Additionally, we propose that using domain-specific objectives (e.g., learning physicochemical properties) during this step will enhance performance due to increased chemical awareness. Employing domain-specific objectives in the DA step can also provide computational efficiency since the size of the DA data is significantly smaller than the pre-training data. To this end, we experimented with two objectives: multi-task regression (MTR) of physicochemical properties, and contrastive learning (CL) of different SMILES representations of the same molecules. In this scheme, the model is pre-trained on a large database of molecules using the MLM objective and then further-trained on the molecules from the downstream tasks using either MLM, MTR, or CL.

Our experiments across seven datasets spanning five ADME endpoints \cite{fang2023prospective, chembl3301361} reveal that increasing the size of the pre-training dataset provided limited benefits, with improvements plateauing at 400 - 800K molecules (30-60\% of the GuacaMol dataset \cite{brown2019guacamol}). In addition, applying domain adaptation with the MTR objective led to significant performance gains across all datasets (P-values < 0.01), an improvement that was not possible by data scaling alone. Furthermore, our transformer models achieved competitive or superior predictive accuracy compared to existing large scale transformer models(MolFormer \cite{ross2022large} and MolBERT \cite{fabian2020molecular}), while being more computationally efficient. Notably, models that incorporate chemically informed features and objectives demonstrated the strongest performance, both among transformers and baselines, highlighting the importance of domain-aware training strategies.

\section{Datasets and Preprocessing}

In this work, we pre-train a transformer model on a dataset sampled from a large-scale general-purpose library of molecules, namely GuacaMol \cite{brown2019guacamol}. We then evaluate our models on a benchmark dataset that explores ADME properties (Absorption, distribution, metabolism, excretion) \cite{fang2023prospective, chembl3301361}. 
In our approach, datasets are utilized in three stages of model training: pre-training, domain adaptation, and evaluation. The GuacaMol dataset is exclusively used for pre-training, the unlabeled ADME dataset for domain adaptation, and the labeled ADME dataset version for evaluation. In this section, we provide a brief description of each dataset.

\subsection{Pre-training Dataset: GuacaMol}
\label{guacamol_dataset}
We pre-train the models on the GuacaMol dataset \cite{brown2019guacamol}. This dataset is a subset of $\sim$1.3M molecules sampled from the ChEMBL database \cite{gaulton2017chembl}, which contains molecules that have been synthesized and assayed with respect to different biological endpoints. The GuacaMol subset was filtered to benchmark generative models in multiple aspects such as molecular generation validity, or similarity. To this end, the ChEMBL database was filtered so that the selected molecules were different from a holdout set of 10 already marketed drugs.
The filtering process was done by removing all molecules that have a similarity above $0.323$ to these 10 drugs measured as Tanimoto similarity using ECFP4 fingerprints. The molecules were also filtered by size, i.e., to contain between 5 and 100 atoms, and by element, i.e., to be part of this atom list: H, B, C, N, O, F, Si, P, S, Cl, Se, Br, and I. Furthermore, they were standardized, i.e., salts were removed and charges neutralized. 

\subsection{Domain Adaptation and Downstream Datasets: ADME benchmark}
For domain adaptation (DA) and evaluation, we utilize seven datasets spanning five ADME endpoints: lipophilicity, permeability, solubility (two datasets), intrinsic clearance (two datasets), and plasma protein binding. The DA step involves further training of a pre-trained model on smaller, domain-specific datasets (i.e., the investigated endpoints) without labels, allowing the model to adapt to the target distribution. Subsequently, in the molecular property prediction (MPP) task, the same datasets—now with labels—are used for evaluation.

The datasets originate from two sources: Fang \textit{et. al.}. \cite{fang2023prospective} and a ChEMBL collection compiled by AstraZeneca \cite{chembl3301361}. Fang \textit{et. al.}. provide datasets for permeability,  solubility, two different assay measurements for intrinsic clearance, and two different assay measurements for plasma protein binding. Figure~\ref{fig:fang_datasets} shows the distribution of target values in these datasets. We excluded the plasma protein binding datasets from Fang \textit{et. al.}. due to their small sample size (fewer than 200 molecules), which is considered insufficient for drawing robust conclusions in regression modeling \cite{babyak2004you}.

\begin{figure}[!ht]
    \centering
    \includegraphics[width=\linewidth]{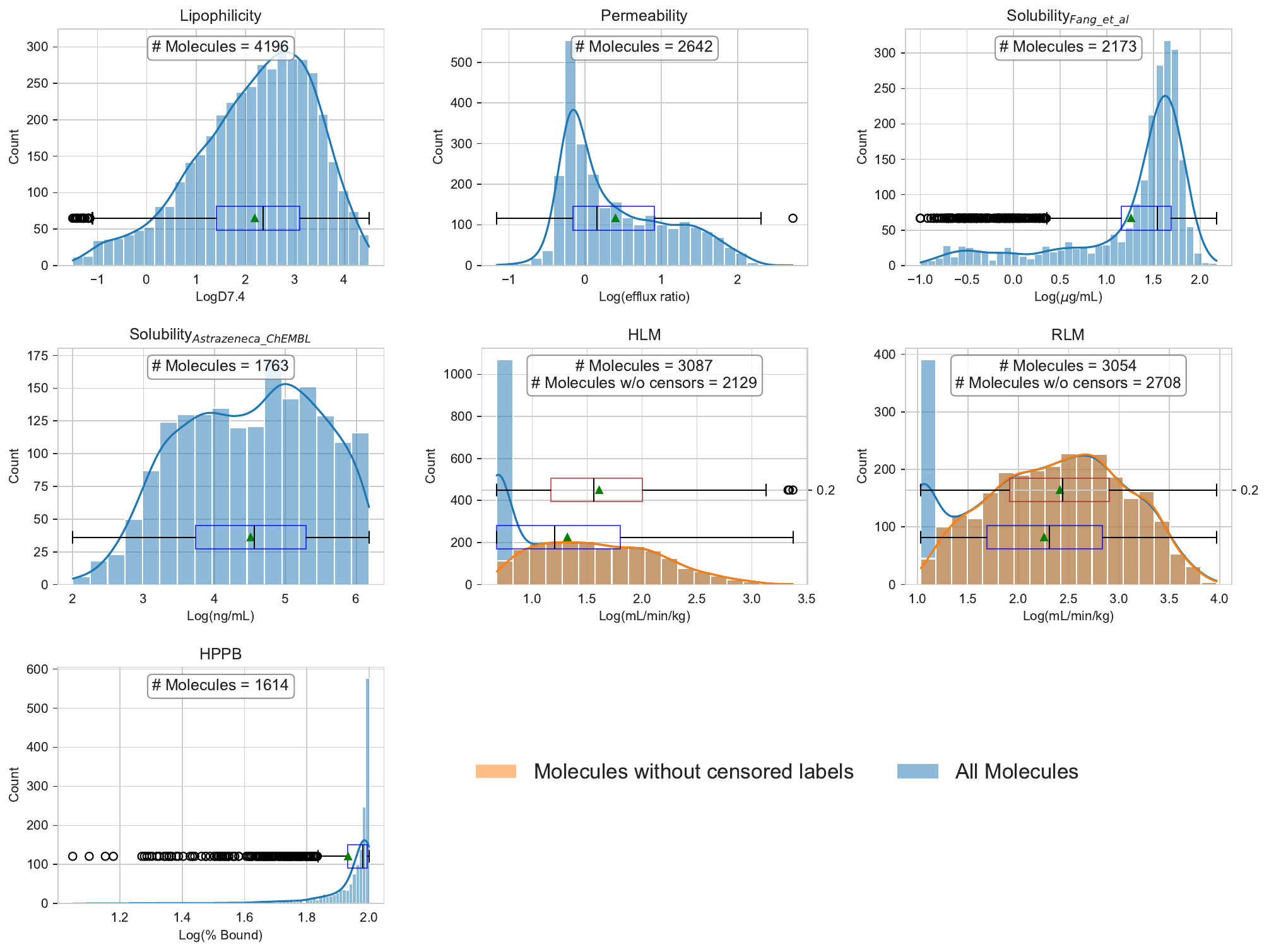}
    \caption{Data summary and distributions of the seven investigated ADME endpoints. The boxplots are shown per histogram to highlight the location of the different quartiles and datapoints that would be seen as outliers. The plots with two distributions show the presence of censored labels which are not suitable for being used directly in a regression model. Therefore, the two distributions show the original data (blue) and the actual data (orange) used during evaluation.}
    \label{fig:adme_summary}
\end{figure}

The AstraZeneca collection encompasses a wider range of ADME endpoints, including lipophilicity, solubility, three different assay measurements for intrinsic clearance, five different plasma protein binding assays, and pKa measurements for the first and second acidic groups, and the first through third basic groups (see Figure~\ref{fig:az_datasets}). From this collection, we include datasets for lipophilicity, solubility, and one plasma protein binding assay --- specifically those containing more than 1,500 molecules, a threshold selected to align with the sample sizes of the Fang \textit{et. al.}. datasets. Refer to Figure \ref{fig:adme_summary} for an overview of the final selected datasets. 
In the following, we detail each endpoint used in this study.

\textbf{Lipophilicity} is a key physicochemical property describing the affinity of a compound for lipophilic (nonpolar) versus hydrophilic (polar) environments, often expressed as a distribution coefficient (logD) or partition coefficient (logP) \cite{Kerns2008}. It plays a critical role in drug discovery by influencing absorption, distribution, metabolism, excretion, and toxicity (ADMET) properties \cite{waring2009defining}. Experimental determination of lipophilicity is commonly performed using the shake flask method, particularly for logD at physiological pH of 7.4. In the dataset sourced from ChEMBL (CHEMBL3301361) \cite{chembl3301361}, lipophilicity data were measured for $\sim$4K molecules by AstraZeneca using the octan-1-ol/water shake flask method, as described by Wenlock \textit{et. al.}. (2011), covering a logD range of -1.5 to 4.5 \cite{wenlock2011method}. 

\textbf{Permeability} assays determine the rate at which a molecule passes through a biological membrane by measuring the ratio between movement to and from the bloodstream \cite{varma2006functional, yang2023current}. Fang \textit{et. al.} uses the MDR1 transfected MDCK cell lines to measure this efflux ratio \cite{fang2023prospective}. MDR1 (also known as P-glycoprotein or P-gp receptor) is present in various tissues, such as intestines and blood-brain barrier, making it a good candidate to assess permeability \cite{wang2005evaluation, varma2006functional, yang2023current}. Figure \ref{fig:adme_summary} shows the distribution of these $\sim2.6$K molecule's efflux ratio.

\textbf{Solubility} is a vital property for understanding the uptake and distribution of a molecule within living organisms, which determines a molecule's efficacy and usability \cite{lipinski2012experimental}. Solubility assays inform about maximum amount of a molecule to dissolve in a solution, either water or a physiologically relevant medium, under certain pH and temperature \cite{barrett2022discovery}. Fang \textit{et. al.}. \cite{fang2023prospective} measured the solubility of $\sim 2.2$K molecules in phosphate buffered saline (PBS) at pH 6.8 following the protocol in Kestranek \textit{et. al.}. \cite{kestranek2013chemiluminescent}. The reported measuring unit is $\mu g/mL$. Figure~\ref{fig:adme_summary} shows the distribution of the measurements. The data has a strong skew (i.e., imbalance) at $\sim 1.5$ log($\mu g/mL$) ($\sim 31.62$ $\mu g/mL$). The second solubility dataset, measured by AstraZeneca, is obtained from ChEMBL \cite{chembl3301361}. In this assay, solubility of $\sim 1.8$K molecules was determined in pH 7.4 buffer using solid starting material, following the shake flask method described by Bevan and Lloyd \cite{wenlock2011highly}. The experimental concentrations range from 2 to 6 log(nM) (i.e., 0.10 to 1500~$\mu$M). The data is quite uniform for the range 3.5 - 6.2, and underrepresented for the values from 2 - 3.5.

\textbf{Microsomal stability} is used to measure the clearance of a molecule from the body, specifically, intrinsic clearance $CL_{int}$ \cite{grasela2017human}. Intrinsic clearance quantifies the volume of incubation medium from which a molecule is cleared per time unit per microsomal weight unit. In Fang et. al. \cite{fang2023prospective}, the measuring unit is mL/min/kg presented for two datasets: human liver microsomes (HLM) and rat liver microsomes (RLM). Figure \ref{fig:adme_summary} shows the data summary of the two datasets with HLM spanning around three log units of mL/min/kg (0.6 - 3.4), while RLM spans around four log units (1 - 4). The HLM dataset has a sharp peak at $\sim 0.67$ ($\sim 4.7$ mL/min/kg). However, further inspection shows that $\sim 27\%$ of the data has the same target value. This observation corresponds to what is called "censored labels", where experimental values can be recorded only up to a certain threshold, and every observation beyond this threshold is set to a constant value \cite{svensson2024enhancing}. When training a regression model to predict continuous values, this censoring is introducing noise to the model. The same censoring problem is seen for the RLM dataset, although less severely as only $\sim 9$\% of the data is censored. While the censored regression problem has been investigated in fields like survival analysis, its resolution in MPP is still in progress \cite{svensson2024enhancing}. In this work, we handle the censored label by removing the molecules with censored labels from the evaluation step to reduce the noise. However, we keep the molecules during the DA step as they still provide information for further training. After removing these values, the number of labeled molecules becomes $\sim 2.1$K and $\sim 2.7$K for HLM and RLM datasets, respectively. The HLM dataset now spans the range 0.7 to 3 log units while RLM spans the same range (1 to 4 log units).

\textbf{Plasma protein binding} assays measure the amount of a compound that has been captured by the plasma proteins in the blood. This assay is important because the percentage of free molecules in the blood is partly responsible for a compound's activity and efficacy. Therefore, knowing the percentage of free molecules can also guide other assays like permeability and clearance \cite{stoner2011pharmacokinetics}. A human plasma protein binding (PPB) dataset from AstraZeneca, available in ChEMBL, was measured using equilibrium dialysis. In this assay, compounds are incubated with whole human plasma at 37$^\circ$C for over 5 hours to reach binding equilibrium. The method follows the protocol described by Testa \textit{et. al.} \cite{testa2006pharmacokinetic}. For $\sim1,600$ molecules, experimental binding values range from 10\% to 99.95\%, equivalent to 0–2 log units. However, the data distribution shows that $\sim$ 70\% of the data is concentrated in 0.06 log units (between 70\% and 100\% bound).

\section{Methods}

\begin{figure}[!ht]
    \centering
    \includegraphics[width=0.8\linewidth]{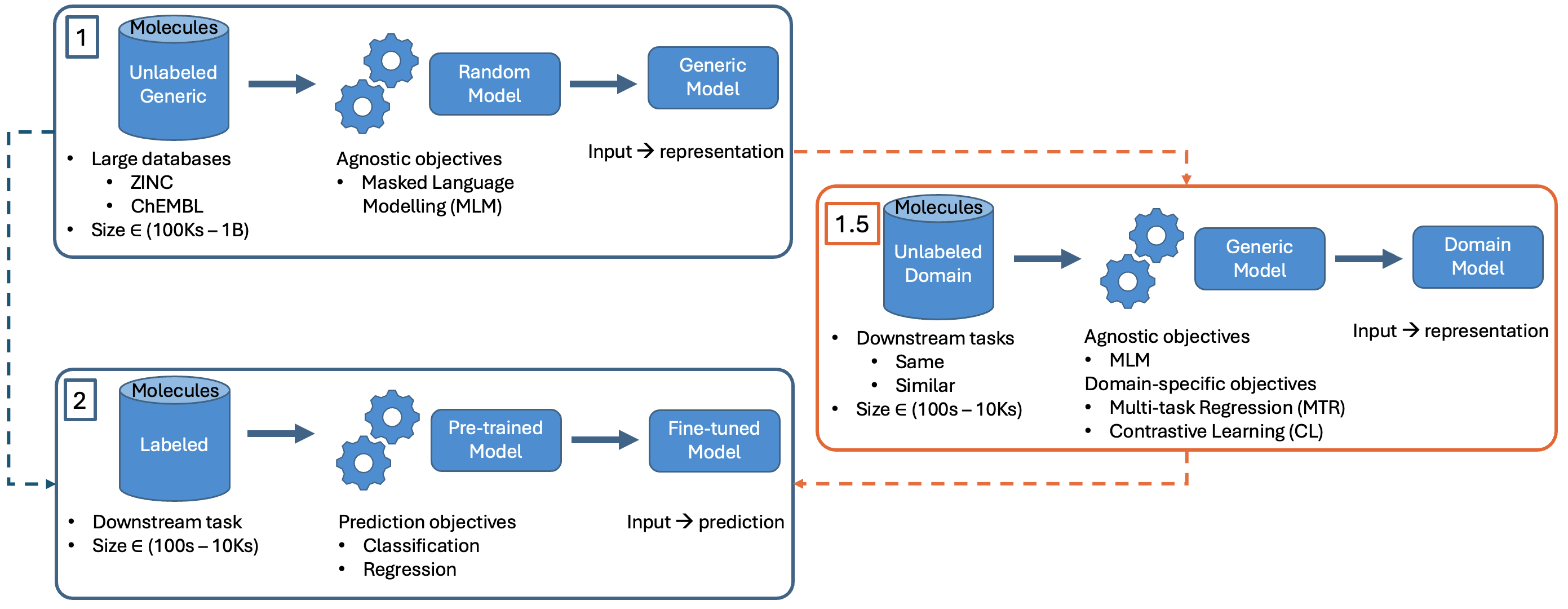}
    \caption{An overview of this research's workflow. Transformer models are trained by pre-training on generic large unlabeled datasets using one or more objectives (step 1), followed by fine-tuning on labeled datasets (step 2). Domain adaptation is an optional intermediate step that resembles pre-training, but can be done on much smaller unlabeled dataset (step 1.5). }
    \label{fig:workflow}
\end{figure}

In this work, we trained a BERT-like model with the simple MLM objective using the GuacaMol \cite{brown2019guacamol} dataset followed by domain adaptation on the ADME benchmark \cite{fang2023prospective} using either MLM or a domain-specific objective, i.e., MTR and CL (Figure \ref{fig:workflow}). We trained multiple models using different sizes of pre-training datasets to further assess the relationship between pre-training dataset size and downstream performance. In the following, we will detail the architecture and training of our models. We describe each training objective, explain the method for increasing the pre-training dataset size, list the literature models used for benchmarking, and finally, outline the evaluation process. Code, data, and analysis can be found at our github repo \url{https://github.com/uds-lsv/domain-adaptation-molecular-transformers}

\subsection{Model Architecture and Training}

In this work, we use the SMILES representations of the molecules as input to our models. We used the SMILES strings as provided by both publications without further standardization \cite{brown2019guacamol, fang2023prospective}. For the transformer model, we use the same architecture as the original BERT \cite{devlin-etal-2019-bert} model with 12 layers, 12 attention heads, and 768-dimensional embeddings, which yields $\sim 89M$ parameters. We use a maximum sequence length of 128 tokens and a vocabulary size of 4096 tokens. We also used absolute positional embeddings and the same tokenizer (i.e., WordPiece \cite{schuster2012japanese, wu2016google}) as BERT. We trained in mixed precision using bf16.

For pre-training, we use the MLM objective (explained below) with a batch size of 16 for 20 epochs (similar to MolBERT \cite{fabian2020molecular}). We used Adam \cite{kingmaAdamMethodStochastic2017} with a learning rate of $3e^{-5}$ and a linear scheduler with 10\% warm-up. The same hyperparameters were used for domain adaptation, however, the objectives varied between MLM, MTR, and CL (explained below).  

For evaluation, we used either the pre-trained model or the domain-adapted model to generate fixed embeddings using the CLS token (i.e., the first token fixed in front of any sequence) for each molecule in the test set. These embeddings were used as representations of the molecules and were used as an input for a random forest (RF) regressor model from sklearn \cite{pedregosa2011scikit} (default settings were used).

\subsection{Training objectives}

The transformer models learn by optimizing a self-supervised training objective like masked language modeling (MLM). In this work, we used MLM as an objective for pre-training, and explored it for domain adaptation as well. Multi-task regression (MTR) was used for domain adaptation and as a pre-training objective for one of the experiments, while contrastive learning (CL) was used for domain adaptation only. Below, we briefly explain each of these objectives. For all objectives, we used a batch size of 16 with a learning rate of $3e^{-5}$ for 20 epochs. The maximum input length was 128 and mean pooling of the tokens was applied to extract a molecule's representation. 

\paragraph{Masked Language Modeling (MLM)} In the MLM objective, 15\% of the input SMILES tokens are randomly masked in each molecule. The language model then outputs the probability distribution over all possible tokens in the vocabulary for each masked token. MLM is quantified by minimizing the cross-entropy loss:

\begin{equation}
\label{cross_ent}
L_{MLM} = -\sum_{c=1}^My_{o,c}\log(p_{o,c}) 
\end{equation} 

Where $M$ corresponds to the total number of classes (i.e., the vocab size), $y_{o,c}$ is the true class of token o, and $p_{o,c}$ is the predicted class.

\paragraph{Multi-Task Regression (MTR)} The MTR objective is independent of the linguistic structure of the SMILES. This objective involves the simultaneous prediction of a vector of $210$ real-valued physicochemical properties of the input molecule. In our work, we use the RDKit \cite{rdkit} framework to calculate the physicochemical descriptors of the pre-training dataset molecules, we normalize the values using the standard scaling, and predict properties using 2 layer MLP with relu activation functions and dropout of 0.1 on the first token. The model uses multi-task mean squared error loss  

\begin{equation}
\label{mse}
L_{MTR} = \sum_{i=1}^{N}\sum_{j=1}^{D}(p_{ij}-y_{ij})^2 
\end{equation}

where $D$ is the $210$-dimensional chemical descriptors and $N$ is the number of training samples, $p_{ij}$ is the predicted value and $y_{ij}$ is the true value. For smooth convergence, we use mean and standard deviation to normalize each descriptor as a pre-processing step.

\paragraph{Contrastive Learning (CL)} The CL objective utilizes the nuances of the SMILES sequence since multiple sequences can represent the same molecules, a process called enumeration. A canonical SMILES sequence per molecule can be generated by fixing the starting and branching procedure of linearizing the molecular graph. We employ CL by using the multiple negatives ranking loss \cite{henderson2017efficientnaturallanguageresponse} which takes SMILES triples. Each triple consists of a canonical SMILES, an enumerated sequence of the same molecule, and the SMILES of a random molecule from the dataset as negative example. The encoder is based on a single BERT model, in which the latent SMILES representations of the canonical and enumerated SMILES are pulled together, whereas the latent representation of the negative SMILES are pushed away from the latent representations of the other two SMILES simultaneously. The contrastive loss is expressed as follows:

\begin{equation}
\label{contrastive_loss_cbert}
    L(r_c, r_e, r_n) = \frac{1}{K} \sum_{i=1}^K \left[ {\rm sim}(r_{c}^{(i)}, r_{e}^{(i)}) - \log \sum_{j=1}^K e^{{\rm sim}(r_{c}^{(i)}, r_{n}^{(j)})} \right]
\end{equation}

Where ${\rm sim}(r_1, r_2)$ is the cosine similarity function defined as $\frac{r_1^Tr_2}{\Vert r_1\Vert\Vert r_2\Vert}$, $r_c$ is the latent representation of the canonical SMILES, $r_e$ is the latent representation of the enumerated SMILES, $r_n$ is the latent representation of the negative SMILES, and$K$ is the number of SMILES triples in the mini-batch.

\subsection{Pre-training dataset selection}
Besides training a model on the full training set of the GuacaMol \cite{brown2019guacamol} dataset (i.e., $\sim 1.3$M molecules), we investigate three more models trained on 0\% (i.e., randomly initialized model with no pre-training), 30\%, and 60\% of the GuacaMol dataset. To select the 30\% and 60\% subsets, the data was first clustered using BitBirch \cite{perez2024efficient} with default settings, and tanimoto similarity based on Morgan fingerprints. Selections were made proportionally from each cluster to preserve the overall cluster distribution across subsets. RDKit \cite{rdkit} was used to calculate fingerprints and similarities, and the selection procedure was adopted from talktorial T005 from TeachOpenCADD \cite{sydow2019teachopencadd}.

\subsection{Dataset splitting and cross validation}
For pre-training, we used the predefined split of the GuacaMol dataset, with an 80:20 ratio for training and testing. During domain adaptation, all molecules from the individual ADME datasets were used for further training without labels. For evaluation on labeled data, we followed the recommendations of Ash \textit{et. al.}. \cite{ash2024practically}, performing $5 \times 5$ repeated cross-validation using both random splitting and cluster-based Butina splitting with Morgan fingerprints (as used in Ash \textit{et. al.}. This cross-validation strategy is designed to produce a large number of independent estimates of the evaluation metric. According to the central limit theorem (CLT), the average of many such estimates tends to follow a normal distribution, enabling more reliable statistical analysis. Satisfying the CLT assumptions allows for the application of parametric statistical tests to assess confidence intervals and significance. In this work, we primarily report results obtained using Butina cluster splitting and refer to results from random splitting where relevant.

\subsection{Evaluation metrics}
For rigorous evaluation, we again follow the recommendations of Ash \textit{et. al.}. \cite{ash2024practically} by examining both error metrics --- like Mean Absolute Error (MAE) and Root Mean Squared Error (RMSE) --- and correlation metrics, including the coefficient of determination ($R^2$) and Pearson correlation ($\rho$). Error metrics assess the model’s accuracy in predicting absolute values, while correlation metrics evaluate its ability to capture trends within the data distribution. Ash \textit{et. al.}. also advocate for parametric significance testing using repeated measures ANOVA (ANOVA-RM) followed by the Tukey Honestly Significant Difference (TukeyHSD) test. This approach accounts for the repeated-measures setup typical in our domain (i.e., evaluating different models on the same test set) and corrects for multiple comparisons across more than two models. The authors of Ash \textit{et. al.} adapted the TukeyHSD implementation to use the standard error of the means derived from ANOVA-RM, rather than from a standard ANOVA. For pairwise model comparisons, we report statistical significance using paired t-tests. In the main manuscript, we primarily report MAE and $R^2$ as our core evaluation metrics; however, additional metrics such as RMSE, $\rho$, and Spearman correlation are available on our GitHub repository.

\subsection{Models benchmarking}
To assess the strengths and limitations of our transformer model, we compare its performance on downstream molecular property prediction tasks using embeddings extracted from our model, two state-of-the-art transformer models --- MolBERT \cite{fabian2020molecular} and MolFormer \cite{ross2022large} --- and molecular features, i.e., physicochemical descriptors and Morgan fingerprints (default settings) computed using RDKit \cite{rdkit}.

We used the original MolBERT implementation \cite{fabian2020molecular} and the HuggingFace implementation of MolFormer \cite{wolf2019huggingface} to extract embeddings. A Random Forest (RF) model with default settings from scikit-learn \cite{pedregosa2011scikit} was trained to predict molecular properties based on each representation: our model's embeddings, MolBERT embeddings, MolFormer embeddings, and molecular features.

Table~\ref{tab:benchmark} provides an overview of each model, including pre-training dataset size, training objectives, domain adaptation details, and the number of parameters. The baseline models serve as simple, transparent benchmarks for comparison against the more complex transformer models.

\begin{table}[!ht]
    \centering
    \begin{tabular}{|l|l|l|l|l|l|}
    \hline
        Model & \multicolumn{2}{|c|}{Pre-training}  & \multicolumn{2}{|c|}{Domain Adaptation}  & \# Parameters  \\ \hline
        ~ & \# Molecules ($\sim$) & Objectives & \# Molecules ($\sim$) & Objectives & ~ \\ \hline
        MLM\_MTR (Ours) & 0, 400K, & MLM or MTR & 170 – 3K & MLM, MTR, & $\sim$89M  \\ 
        ~ & 800K, 1.3M &~ & ~ & or CL & ~  \\ \hline
        MolBERT & 1.3M  & MLM, MTR, & - & - & $\sim$85M \\ 
         & & SMILES-Eq & & &  \\ \hline
        MolFormer & 100M & MLM & - & - & $\sim$85M \\ \hline
        RF + physchem & - & - & - & - & - \\ \hline
        RF + Morgan fingerprint & - & - & - & - & - \\ \hline
    \end{tabular}
    \caption{An overview of the benchmarked models explaining their set-up in terms of pre-training dataset size, objectives and number of trainable parameters. MLM: masked language modeling, MTR: multi-task regression of physicochemical properties, SMILES-Eq: predicting whether two SMILES strings correspond to the same molecule, and CL: contrastive learning of the different representations of the SMILES string.}
    \label{tab:benchmark}
\end{table}

\section{Results and discussion}

In this section, we discuss our findings on the effect of increasing pre-training dataset size, the benefit of using domain adaptation (DA), and how the most efficient and performant setup compares to models from the literature. 

\subsection{Pre-training improves performance on downstream endpoints up to a limit}
Frist, we compare the model performances with varying pre-training dataset sizes, i.e, 0\%, 30\% ($\sim 400K$), 60\% ($\sim 800K$), and 100\% ($\sim 1.3M$) molecules.

\begin{figure}[H]
    \centering
    \includegraphics[width=\linewidth]{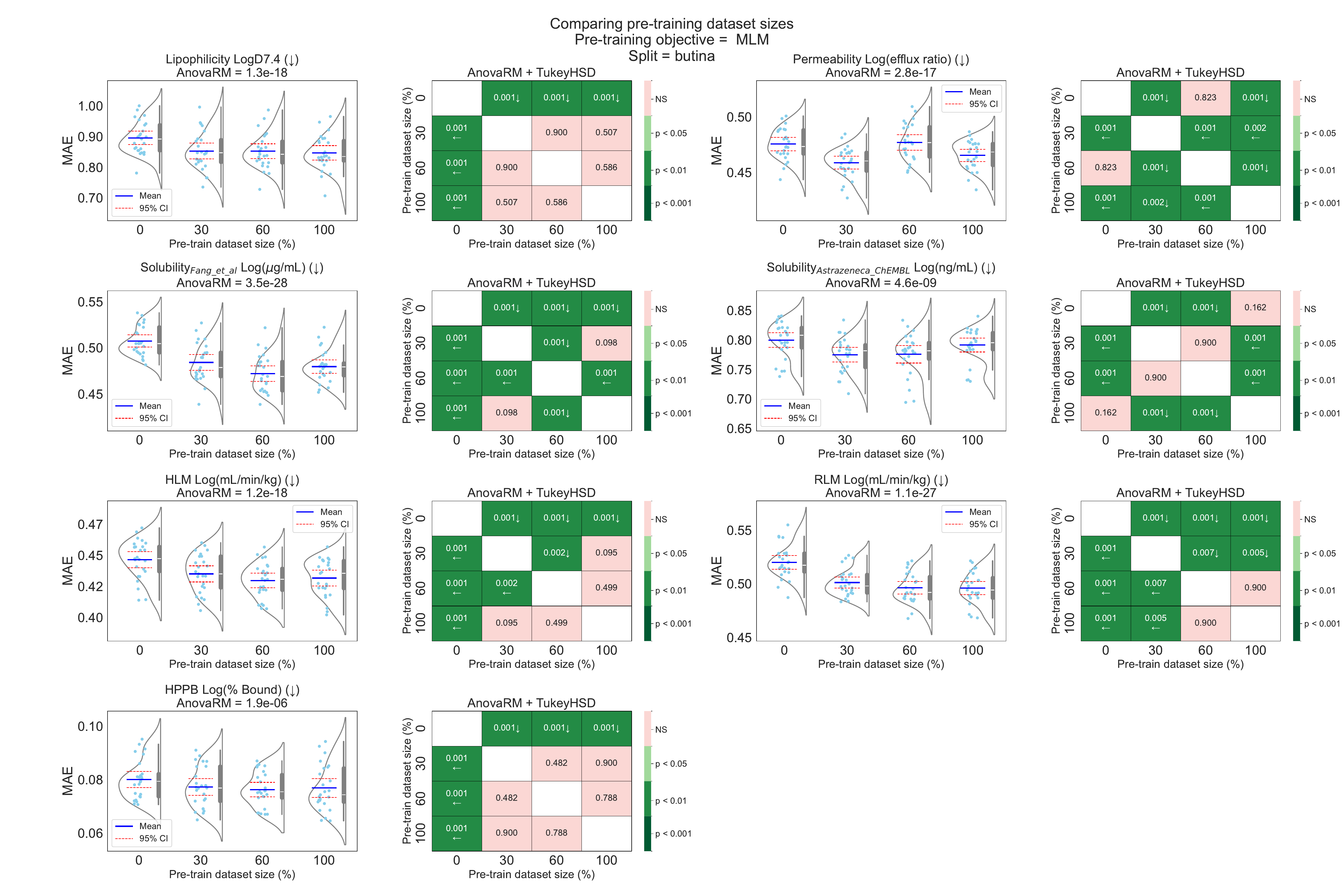}
    \caption{MAE performance for increasing pre-training dataset sizes using Butina splitting. 0\% corresponds to a randomly initialized model with no pre-training and 100\% correspond to the $\sim 1.3$M molecules of the GuacaMol dataset. Two-tailed significance analyses were performed, therefore, the arrows in the heatmap helps recognizing the model with the improved performance. CI = confidence interval for the estimation of the mean.}
    \label{fig:mlm_scaling_mae}
\end{figure}

\begin{figure}[!ht]
    \centering
    \includegraphics[width=\linewidth]{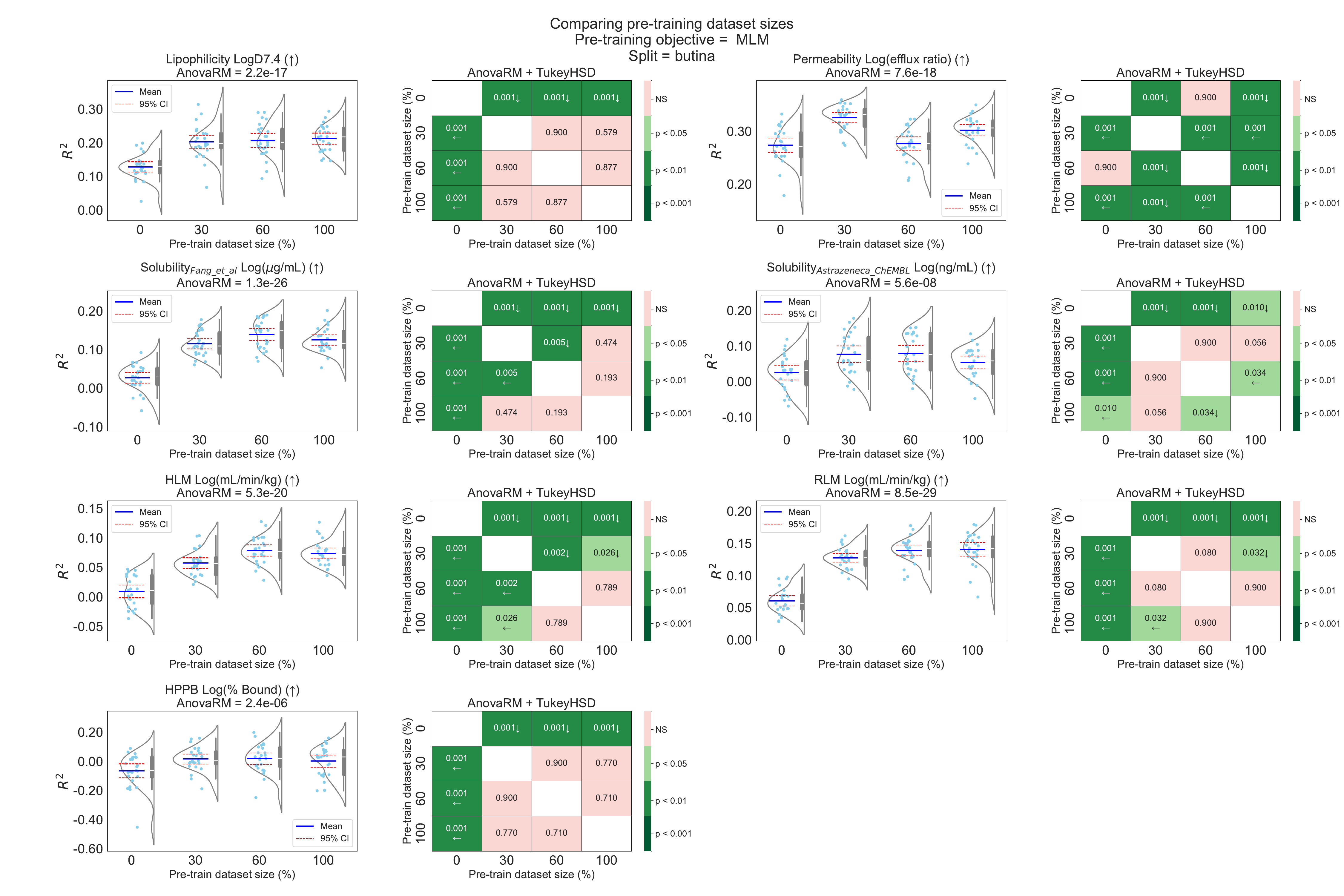}
    \caption{$R^2$ performance for increasing pre-training dataset sizes using Butina splitting. 0\% corresponds to a randomly initialized model with no pre-training and 100\% correspond to the $\sim 1.3$M molecules of the GuacaMol dataset. Two-tailed significance analyses were performed, therefore, the arrows in the heatmap helps recognizing the model with the improved performance. CI = confidence interval for the estimation of the mean.}
    \label{fig:mlm_scaling_R2}
\end{figure}

Figures \ref{fig:mlm_scaling_mae} and \ref{fig:mlm_scaling_R2} show the MAE and $R^2$ performance of these models on seven independent molecular property datasets. The figures show that, for all datasets, at least two pre-training setups provided significantly better predictions than using a randomly initialized model (P-val < 0.01). However, the figures also show that, for all datasets, the 100\% setup was either matched in performance (P-val > 0.05 for lipophilicity, HLM, and HPPB) or outperformed (P-val < 0.01 for permeability, the two solubility datasets, and RLM) by the 30\% or the 60\% setups. This shows that increasing pre-training dataset size plateaus, and further training is not recommended. 

In comparing the 30\% and 60\% setups, significance analysis revealed mixed results, with no consistent advantage for either configuration. For lipophilicity, solubility$_{\text{Astrazeneca\_ChEMBL}}$, and HPPB, no significant differences were observed ($p$-value~$>$~0.4), whereas the 30\% setup performed significantly better for permeability ($p$-value~=~0.001). In contrast, the 60\% setup outperformed the 30\% setup for solubility$_{\text{Fang\_et\_al}}$, HLM, and RLM ($p$-value~$<$~0.01). Given the lack of strong evidence favoring one setup overall, we selected the 30\% configuration for subsequent analyses due to its greater efficiency.

While significance testing assesses whether observed performance differences between models are statistically meaningful, it does not capture the practical relevance of these differences. 
A comprehensive evaluation therefore also requires examining the absolute values of the performance metrics.
Mean Absolute Error (MAE), being unbounded and dataset-dependent, can be difficult to interpret without contextual knowledge. In such cases, experts in the specific application domain are typically better positioned to assess whether an MAE is acceptable.

Figure~\ref{fig:mlm_scaling_R2} shows the results for the $R^2$ metric, which is bounded within $[-\infty, 1]$. Negative $R^2$  values indicate worse performance than simply predicting the mean, and a value of 1 corresponds to perfect prediction. For each dataset, the average $R^2$ in repeated runs remains notably low, for example, 0.08 for Solubility$_{Astrazeneca\_ChEMBL}$, 0.06 for HLM, and as low as 0.02 for HPPB. The highest reported value is 0.33 for Permeability. 
Although statistical significance is observed, the absolute values of a bounded and interpretable metric highlight the remaining challenges in capturing the complex relationship between molecular structure and properties.

These findings highlight the critical need and opportunities for advancing transformer-based architectures in molecular property prediction through means other than dataset scaling alone.

Finally, the results presented in Figures~\ref{fig:mlm_scaling_mae} and~\ref{fig:mlm_scaling_R2} are based on train/test dataset splitting using Butina clustering with Morgan fingerprints, which is generally considered a stricter and more realistic evaluation strategy compared to random splitting. Figures~\ref{fig:adme_mlm_scaling_mae_random} and~\ref{fig:adme_mlm_scaling_r2_random} display the results from random splitting, which paint a similar overall picture to that observed with Butina clustering. While pre-training proves beneficial, performance tends to plateau around the 30\% or 60\% pre-training levels.

Furthermore, although the $R^2$ values under random splitting are generally higher than those observed with Butina clustering --- as expected due to the less stringent data separation, they remain relatively low overall. The highest average $R^2$ value was observed for Lipophilicity, with an $R^2$ of 0.39. 

Achieving comparable results under the more relaxed random splitting strategy further emphasizes the need for more effective approaches to model the structure–property relationship.

\subsection{Domain adaptation improves performance significantly}
\begin{figure}[h]
    \centering
    \includegraphics[width=\linewidth]{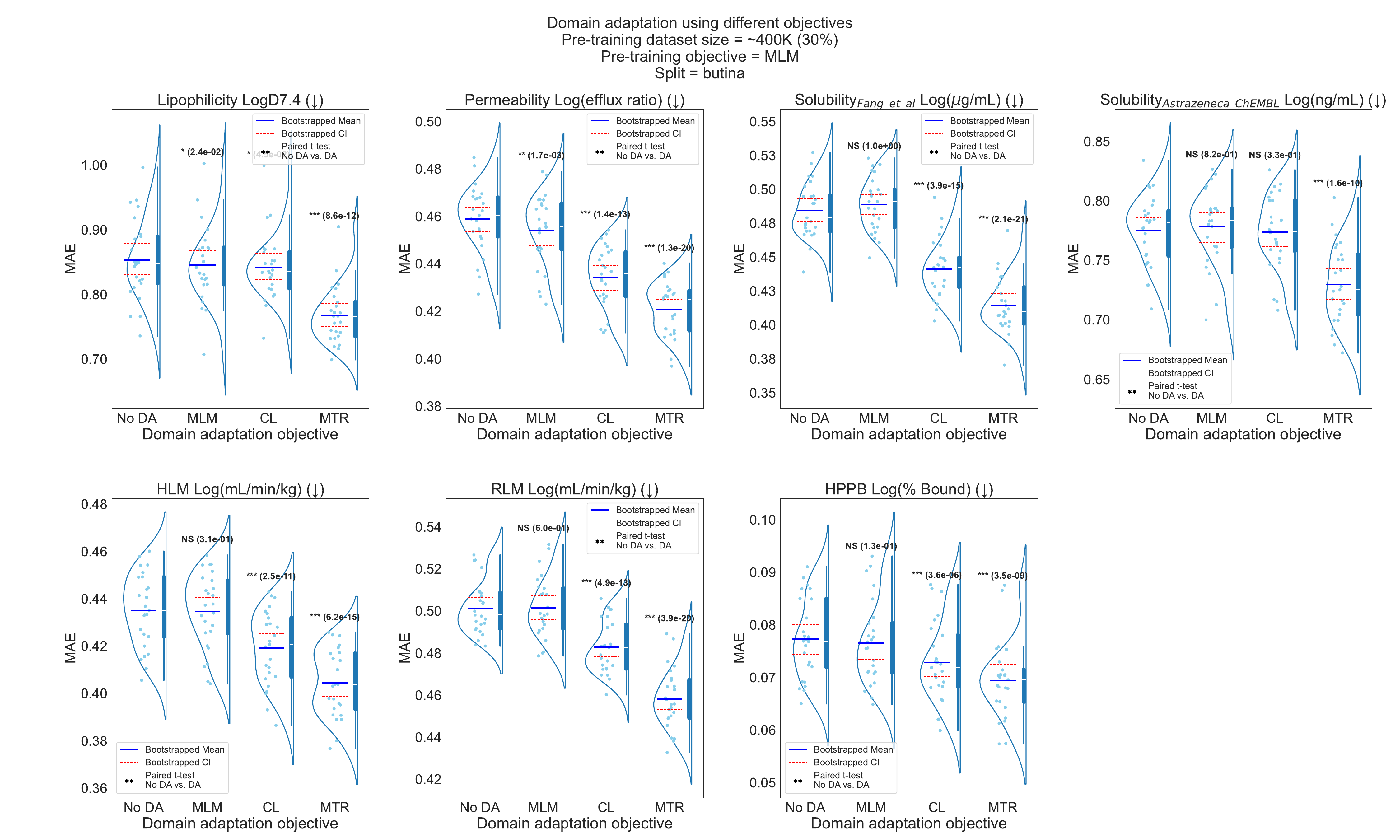}
    \caption{MAE performance for a baseline model trained with pre-training only (No DA), and three models incorporating domain adaptation (DA) using different objectives: Masked Language Modeling (MLM), Contrastive Learning (CL), and Multi-task Regression (MTR) for physicochemical properties. P-values are from one-tailed paired t-tests comparing each DA model to the No DA baseline, under the hypothesis that DA improves performance. Significance levels: * $p < 0.05$, ** $p < 0.01$, *** $p < 0.001$.}
    \label{fig:adme_da_MAE}
\end{figure}

\begin{figure}[h]
    \centering
    \includegraphics[width=\textwidth, height=\textheight, keepaspectratio]{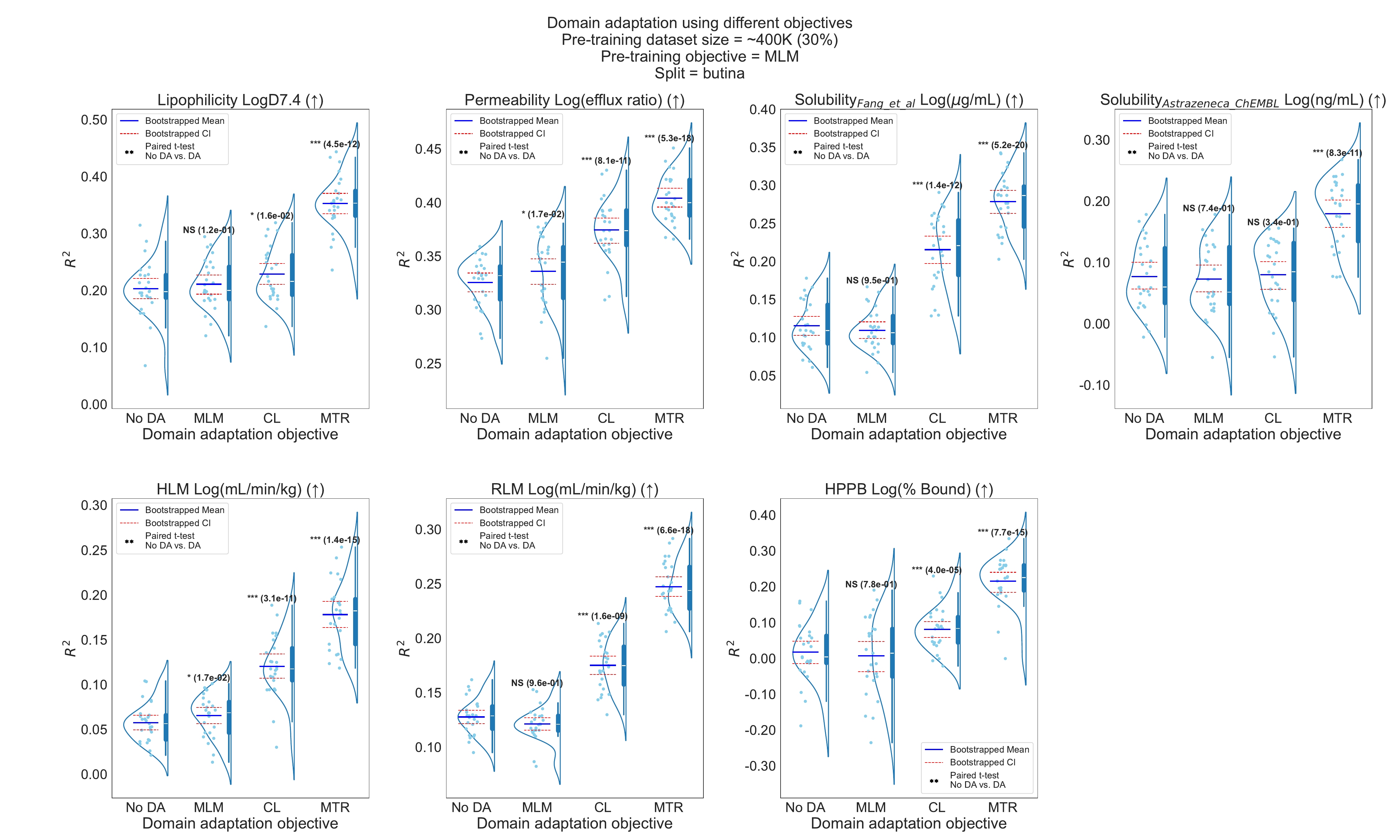}
    \caption{$R^2$ performance for a baseline model trained with pre-training only (No DA), and three models incorporating domain adaptation (DA) using different objectives: Masked Language Modeling (MLM), Contrastive Learning (CL), and Multi-task Regression (MTR) for physicochemical properties. P-values are from one-tailed paired t-tests comparing each DA model to the No DA baseline, under the hypothesis that DA improves performance. Significance levels: * $p < 0.05$, ** $p < 0.01$, *** $p < 0.001$.}
    \label{fig:adme_da_R2}
\end{figure}

For the DA studies, we select the model trained on 30\% of the Guacamol dataset as an efficient baseline and compare domain adaptation using three different objectives: masked language modeling (MLM), contrastive learning (CL), and physicochemical multi-task regression (MTR).
Figure \ref{fig:adme_da_MAE} presents the MAE performance for these four models. The results show that domain adaptation using the MTR objective consistently yields highly significant improvements across all datasets ($p$-value < $4e^{-9}$). The CL objective also demonstrates significant gains for all datasets except Solubility$_{Astrazeneca\_ChEMBL}$. In contrast, the MLM objective outperforms the baseline model without domain adaptation on only two datasets, Lipophilicity and Permeability.
Figure \ref{fig:adme_da_R2} presents the corresponding $R^2$ values, offering a diagnostic view of model performance.  Notably, with domain adaptation using the MTR objective, the average $R^2$ values increased by around 0.2 units with the lowest average at $\sim0.2$ , a substantial improvement over the near-zero values observed without adaptation. However, the maximum $R^2$ remains around 0.4, which is still relatively low and suggests that the model continues to face challenges in fully capturing the underlying variability in the data.
These results demonstrate that chemically informed domain adaptation strategies, particularly those leveraging physicochemical properties, offer a more promising path for improving transformer-based molecular property prediction than scaling data alone.

The same trends are observed under the more relaxed random split, as shown in Figures~\ref{fig:da_mae_random} and~\ref{fig:da_r2_random}. Domain adaptation using the MTR objective again yields consistent and highly significant performance improvements across all datasets. For the random-split scenario, the $R^2$ values increase from 0.2 to as high as 0.6. 

\subsection{MTR is a better pre-training objective, but best used for domain adaptation}

Based on the strong performance of the MTR objective in domain adaptation, we further explored its use during pre-training and tested where it works best --- pre-training only, domain adaptation only, or both.
To this end, we pre-trained the same baseline model using the MTR objective alone, and subsequently applied MTR again for domain adaptation.

\begin{figure}[h]
    \centering
    \includegraphics[width=\linewidth]{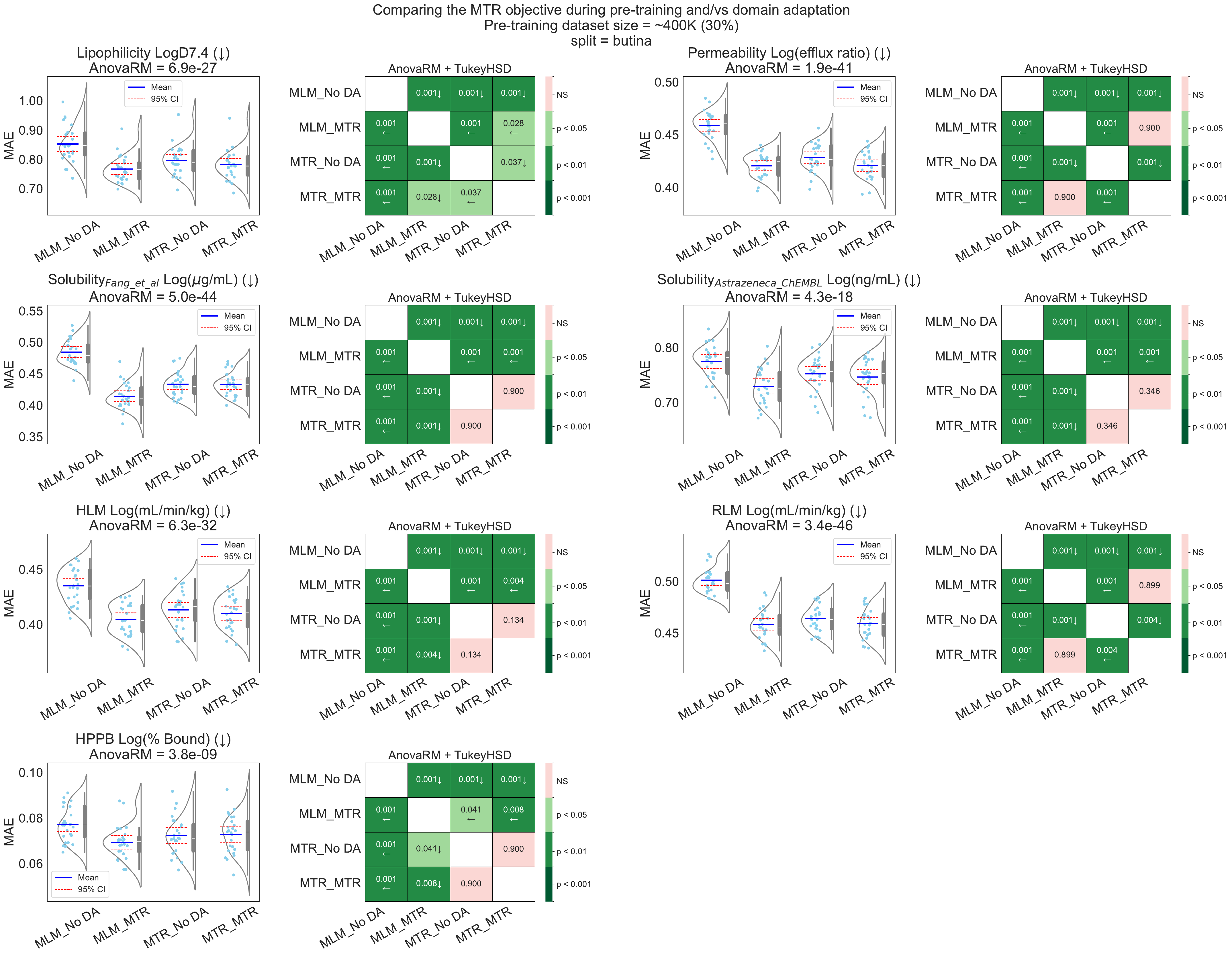}
    \caption{MAE performance to compare the best utilization of the MTR objective in either pre-training only, domain adaptation only, or both. MLM\_No DA = Pre-training with MLM without domain adaptation. MLM\_MTR = Pre-training with MLM and domain adaptation with MTR. MTR\_No DA = Pre-training with MTR only. MTR\_MTR = pre-training with MTR and domain adaptation with MTR. Two-tailed significance analysis were performed, therefore, the arrows in the heatmap helps recognizing the model with the improved performance. CI = confidence interval for the estimation of the mean.} 
    \label{fig:mlm_vs_mtr_butina_MAE}
    
\end{figure}

Figure \ref{fig:mlm_vs_mtr_butina_MAE} compares the performance of four models: one pre-trained with MLM only, one pretrained with MLM and adapted with MTR, one pre-trained with MTR only, and one pre-trained with MTR and adapted with MTR. The results show that pre-training with MTR consistently led to significantly better performance than MLM-based pre-training ($p$-value < 0.01). This highlights the added value of chemically informed objectives over generic, data-agnostic ones like MLM.

Interestingly, the best overall performance was observed for the model pre-trained with MLM and adapted with MTR (denoted as MLM\_MTR). This model achieved significantly better results than the one pre-trained solely with MTR across all datasets except RLM, and also performed better than the model using MTR in both stages for all datasets except Permeability and RLM, where the performances were comparable.

These findings suggest that chemically informed objectives such as MTR are particularly effective when applied to data that is directly relevant to the downstream task. In this case, MTR yielded the greatest benefit when used during domain adaptation, where the model could leverage task-specific information. This underlines the importance of carefully aligning both the dataset and training objective with the prediction target in order to make the most of domain adaptation.

\subsection{Our efficient transformer model rivals more complex models, but not the simplest!}

As examined so far, pre-training on fewer than 400K molecules combined with domain adaptation using the MTR objective on a relatively small number of samples ($\leq$ 4K molecules) yields better performance than extensive pre-training alone on more than 400K molecules. The optimal configuration, referred to as MLM\_MTR, combines MLM for pre-training with MTR for domain adaptation.

We compare the performance of the MLM\_MTR model to two recent transformer-based models: MolBERT~\cite{fabian2020molecular} and MolFormer~\cite{ross2022large}. MolBERT was pre-trained on approximately 1.3 million molecules from the GuacaMol dataset, using a combination of three objectives: MLM, MTR, and SMILES equivalence (SMILES-Eq). MolFormer, on the other hand, was pre-trained on 100 million molecules from the ZINC and PubChem databases, using only the MLM objective. In addition, we include two baseline models based on Random Forest (RF) trained on either physicochemical descriptors or Morgan fingerprints.

\begin{figure}[!ht]
    \centering
    \includegraphics[width=\linewidth]{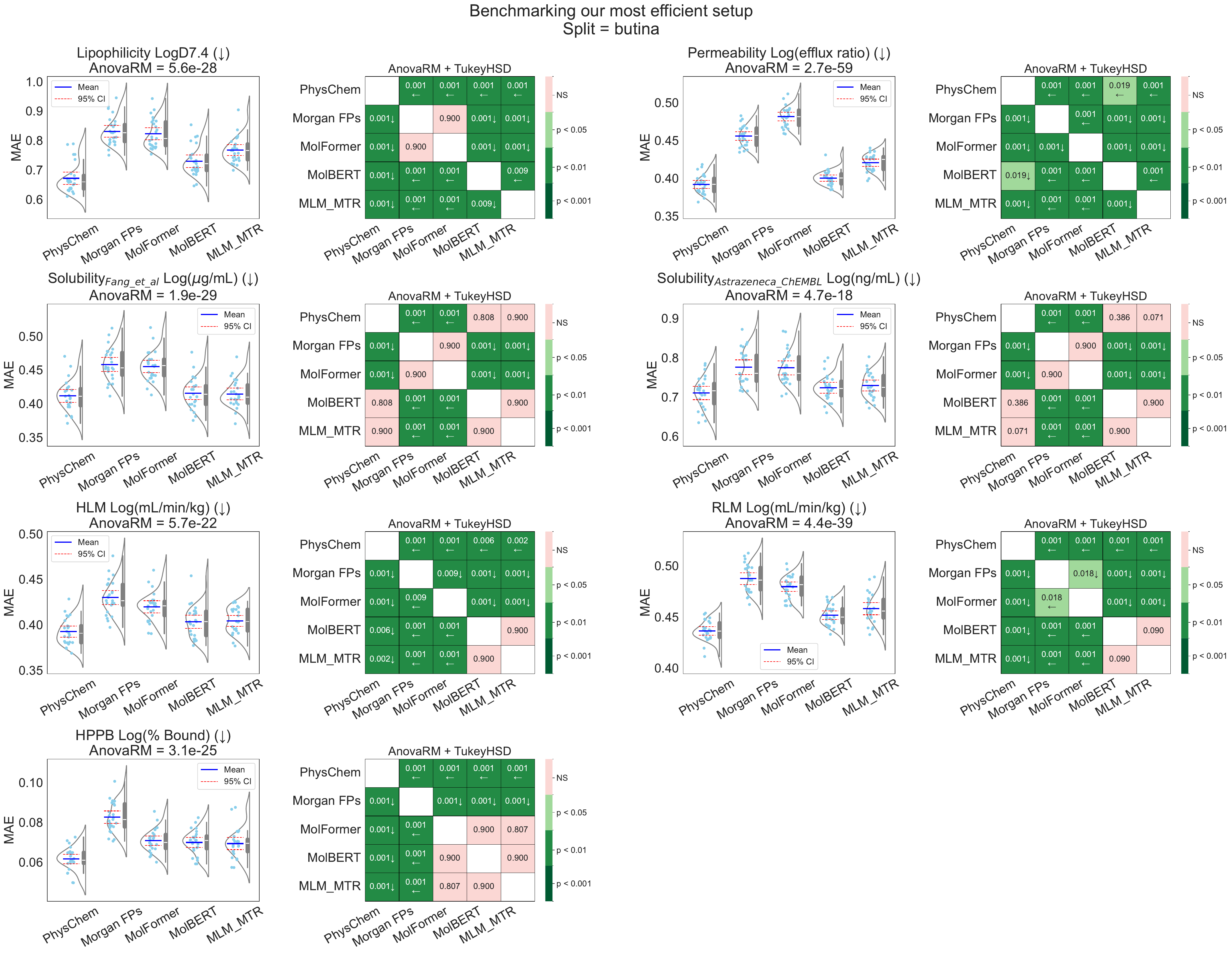}
    \caption{MAE performance of the most efficient model from our analysis to models from the literature. MLM\_MTR corresponds to a transformer model pre-trained with $\sim 400$K molecules using the MLM objective then domain adapted on the corresponding endpoint using the MTR objective. MolBERT \cite{fabian2020molecular} is a chemically aware transformer than has been pre-trained on $\sim 1.3$M molecules using MLM, MTR, and SMILES-EQ objectives. MolFormer \cite{ross2022large} is a large-scale transformer pre-trained on 100M molecules using MLM. PhysChem and Morgan fingerprint correspond to two baselines using Random Forest models. Two-tailed significance analysis were performed, therefore, the arrows in the heatmap helps recognizing the model with the improved performance. CI = confidence interval for the estimation of the mean.}
    \label{fig:adme_comparisons_MAE}
\end{figure}

The results shown in Figure~\ref{fig:adme_comparisons_MAE} indicate that our model outperforms the Morgan fingerprint baseline across all datasets ($p$-value < 0.01), and also surpasses MolFormer on all datasets except HLM ($p$-value < 0.01). MolBERT achieves better performance than our model on only two datasets (Lipophilicity and Permeability), while performance is comparable on the remaining five datasets. These findings suggest that our lightweight model competes effectively with larger-scale models like MolFormer and MolBERT.

Interestingly, the strongest performer overall was the RF model trained on raw physicochemical descriptors, which achieved the best results on five datasets (Lipophilicity, Permeability, HLM, RLM, and HPPB) and performed comparably to MolBERT and our model on the remaining two solubility datasets. However, the $R^2$ values for the strongest model are still around 0.3 for most datasets (Solubility, HLM, RLM, and HPPB), and less than 0.6 for the remaining two datasets, lipophilicity and permeability (Figure \ref{fig:benchmark_r2_butina}. 

This comparison reveals an important trend: the top-performing models in this study (RF + PhysChem, MolBERT, and MLM\_MTR) incorporate explicit physicochemical information. In contrast, structure-based models (using Morgan fingerprints and MolFormer) showed comparatively lower performance. These results reinforce the importance of chemically and physically informed features for molecular property prediction, especially for ADME datasets. Moreover, the superior performance of the RF model on raw physicochemical descriptors, compared to transformer models using the same features as part of their objectives, suggests that these features were not fully exploited within the transformer architectures. This can potentially happen due to the architectural mismatch with low-dimensional, tabular data. While RFs are well-suited for this type of structured input, transformer models, originally designed for high-dimensional, unstructured data, may be over-parameterized and under-optimized for learning from a limited set of descriptors. This aligns with findings from recent studies showing that tree-based models often outperform deep learning models on tabular data without task-specific adaptations~\cite{shwartz2022tabular, gorishniy2021revisiting}.

\section{Conclusion}
In this study, we systematically investigated the performance of transformer-based models for seven molecular property datasets, evaluating the effects of pre-training strategies and domain adaptation objectives. Our results show that while large-scale pre-training with generic objectives like masked language modeling (MLM) offers some benefit, performance plateaus beyond a certain scale. In contrast, domain adaptation using a chemically informed multi-task regression (MTR) objective on domain molecules led to consistent and statistically significant improvements across diverse ADME datasets, even when applied to $\leq$ 4K molecules.

The most effective configuration was a model pre-trained on $\sim400$K molecules (30\% of the GuacaMol dataset) and hybrid objectives that combined MLM for pre-training and MTR for domain adaptation. This model (MLM\_MTR) demonstrated competitive or superior performance to larger transformer models such as MolBERT \cite{fabian2020molecular} and MolFormer \cite{ross2022large}, despite being trained on significantly fewer molecules and lightweight objectives. Furthermore, our comparison with traditional machine learning baselines revealed that models explicitly leveraging physicochemical descriptors—like Random Forests or MTR-adapted transformers—outperformed purely structure-based approaches.

While the baseline Random Forest model using raw physicochemical properties remained the strongest overall, our findings highlight clear and practical strategies for enhancing the performance of transformer models. Specifically, incorporating chemically informed objectives, aligning model adaptation with task-relevant data, and using domain adaptation in addition to pre-training were all critical factors.

\section{Data and code availability}

All data, analysis and implementation code can be found in our github repository at \url{https://github.com/uds-lsv/domain-adaptation-molecular-transformers}. The models pre-trained with 30\% and 60\% using MLM or MTR can be found on our HuggingFace collection at \url{https://huggingface.co/collections/UdS-LSV/domain-adaptation-molecular-transformers-6821e7189ada6b7d0a5b62d4}. We additionally uploaded the domain-adapted models using MTR for each examined endpoint.

\bibliographystyle{unsrt}
\bibliography{references}

\clearpage  
\section*{Supporting Information}  

\setcounter{figure}{0}
\renewcommand{\thefigure}{S\arabic{figure}}

\setcounter{table}{0}
\renewcommand{\thetable}{S\arabic{table}}

\begin{figure}[H]
    \centering
    \includegraphics[width=0.8\linewidth]{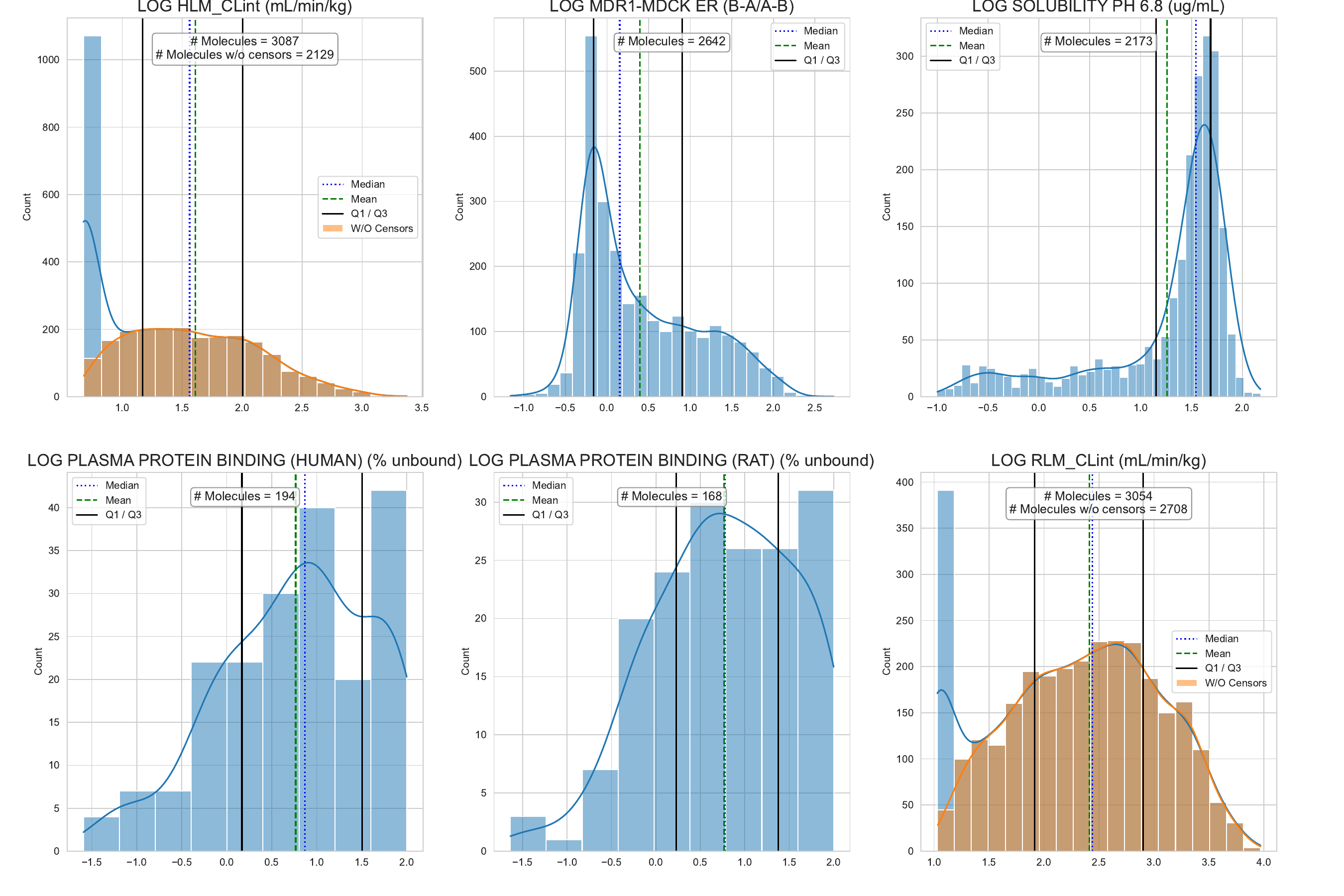}
    \caption{Fang \textit{et. al.} \cite{fang2023prospective} dataset summary and distribution.}
    \label{fig:fang_datasets}
\end{figure}

\begin{figure}
    \centering
    \includegraphics[width=0.8\linewidth]{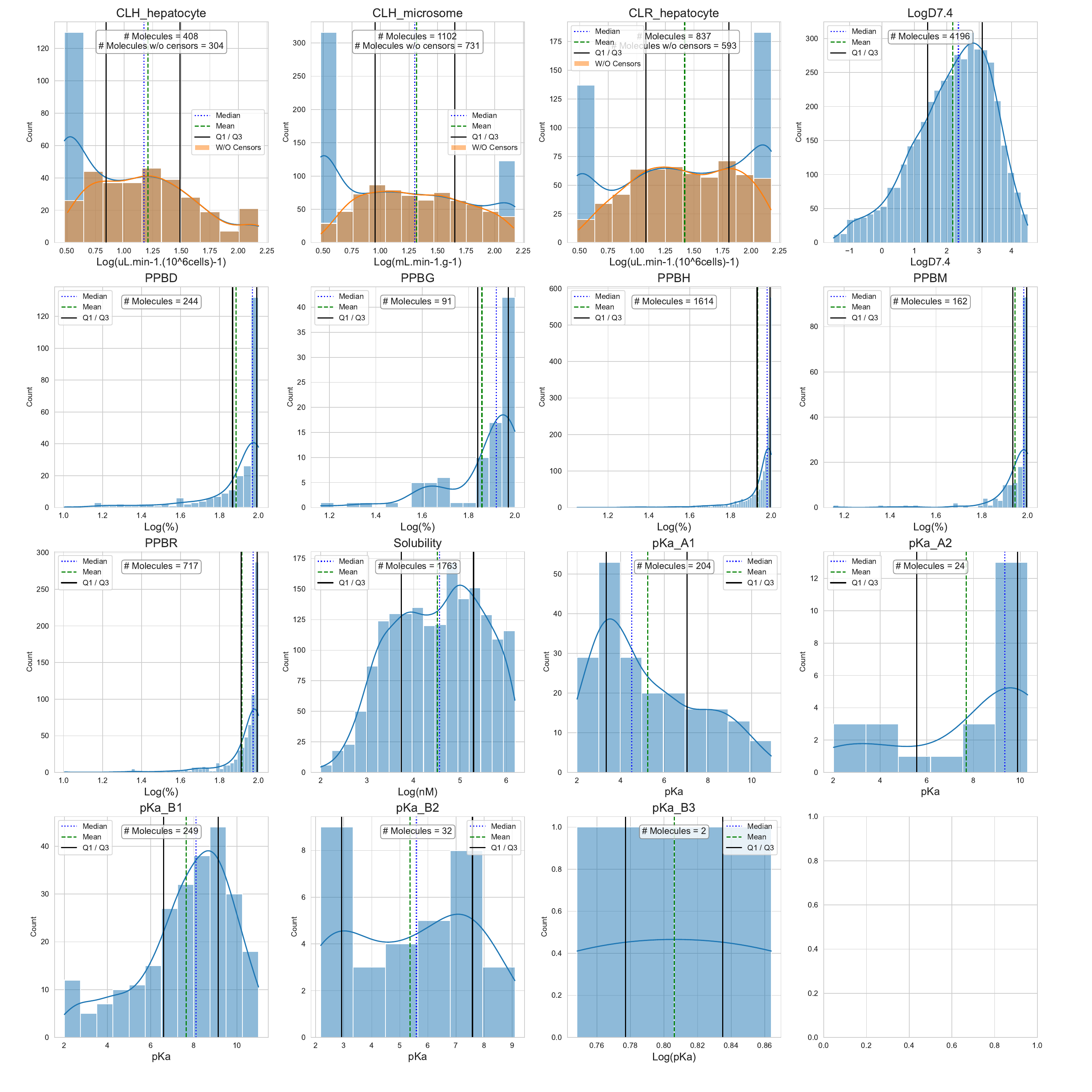}
    \caption{AstraZeneca ChEMBL \cite{chembl3301361} dataset summary and distribution.}
    \label{fig:az_datasets}
\end{figure}

\begin{figure}
    \centering
    \includegraphics[width=0.8\linewidth]{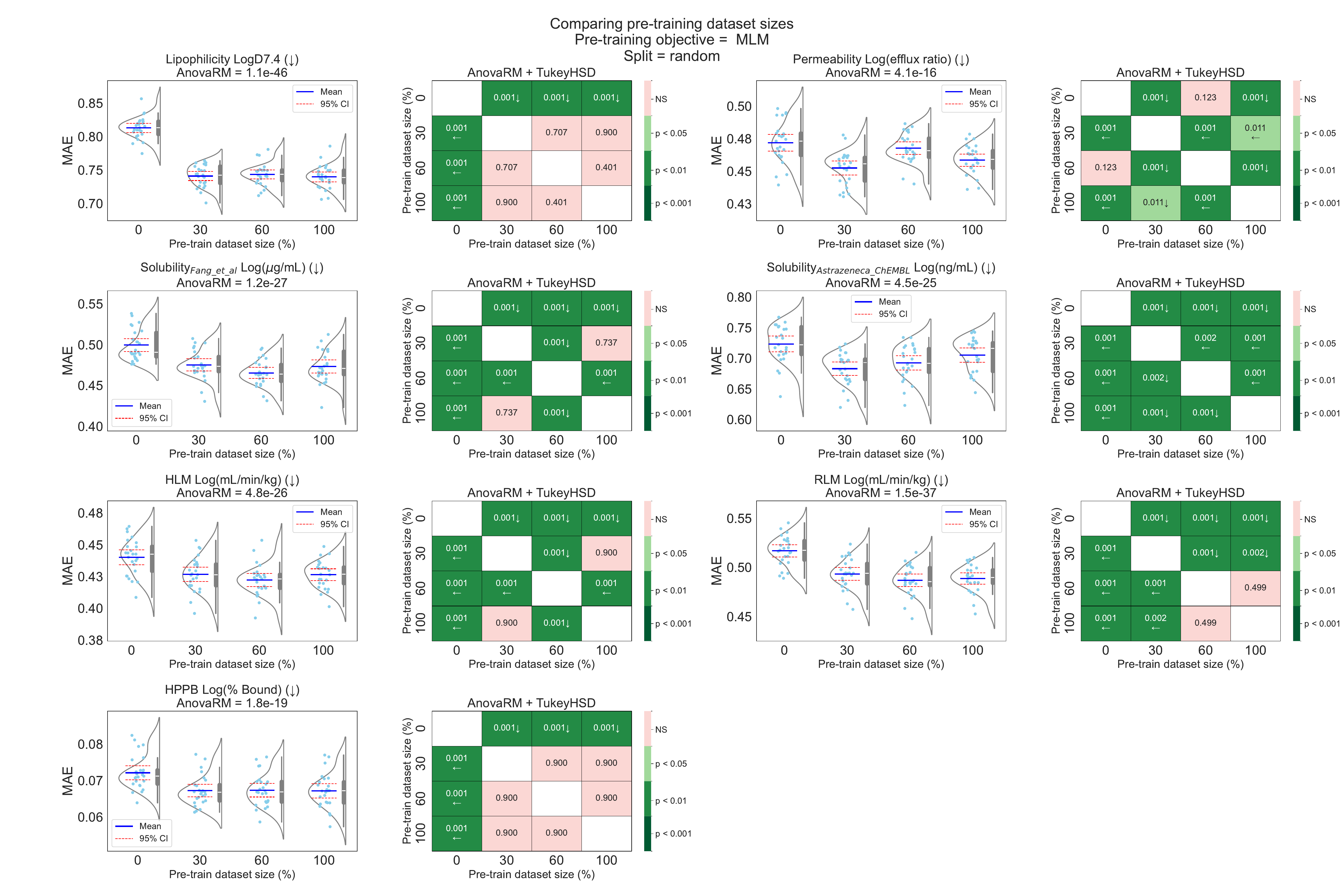}
    \caption{MAE performance for increasing pre-training dataset size using random splitting. 0\% corresponds to a randomly initialized model with no pre-training and 100\% correspond to the $\sim$1.3M molecules of the GuacaMol dataset. Two-tailed significance analysis were performed, therefore, the arrows in the heatmap helps recognizing the model with the improved performance. CI = confidence interval for the estimation of the mean.}
    \label{fig:adme_mlm_scaling_mae_random}
\end{figure}

\begin{figure}
    \centering
    \includegraphics[width=0.99\linewidth]{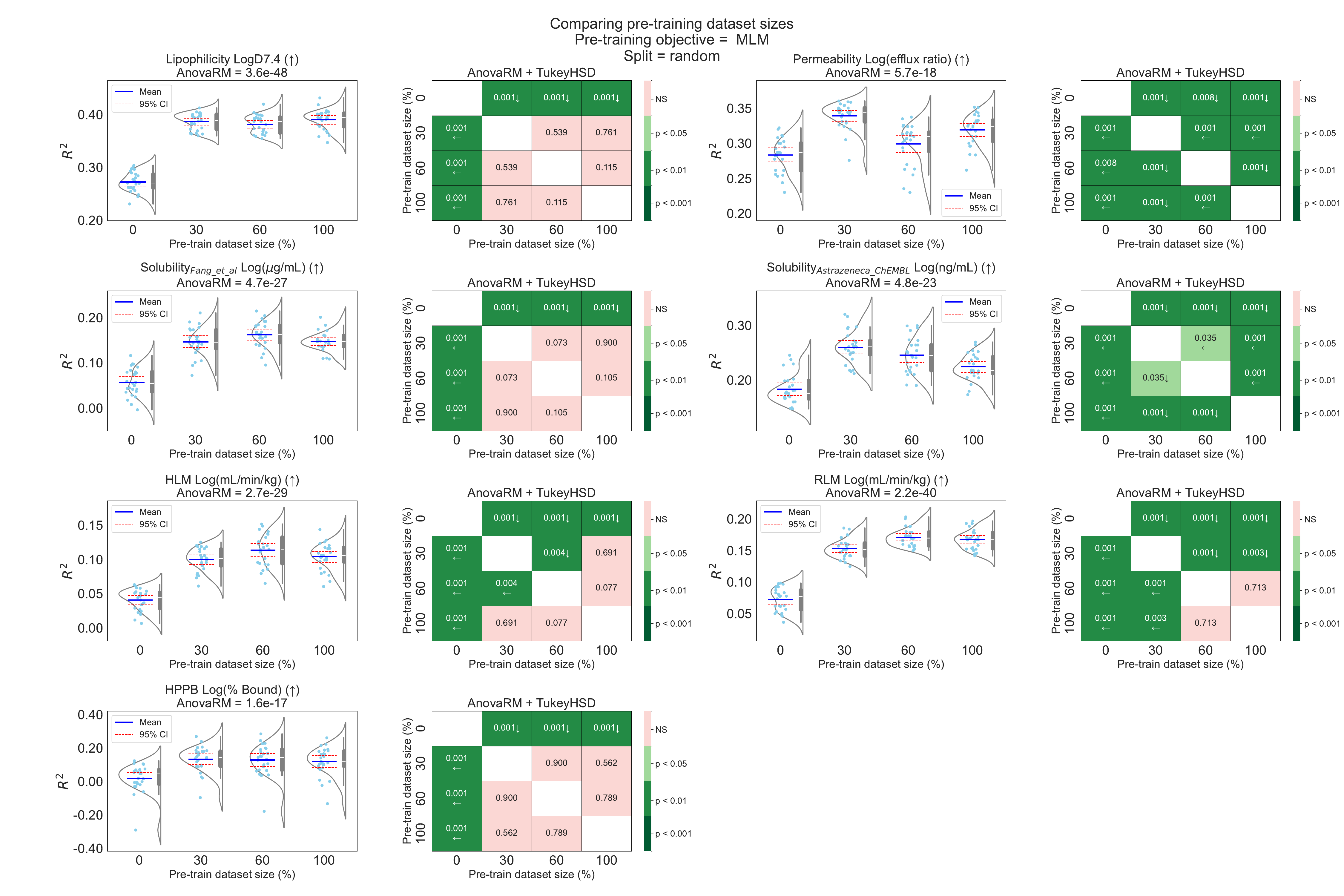}
    \caption{$R^2$ performance for increasing pre-training dataset size using random splitting. 0\% corresponds to a randomly initialized model with no pre-training and 100\% correspond to the $\sim$1.3M molecules of the GuacaMol dataset. Two-tailed significance analysis were performed, therefore, the arrows in the heatmap helps recognizing the model with the improved performance. CI = confidence interval for the estimation of the mean.}
    \label{fig:adme_mlm_scaling_r2_random}
\end{figure}

\begin{figure}
    \centering
    \includegraphics[width=0.8\linewidth]{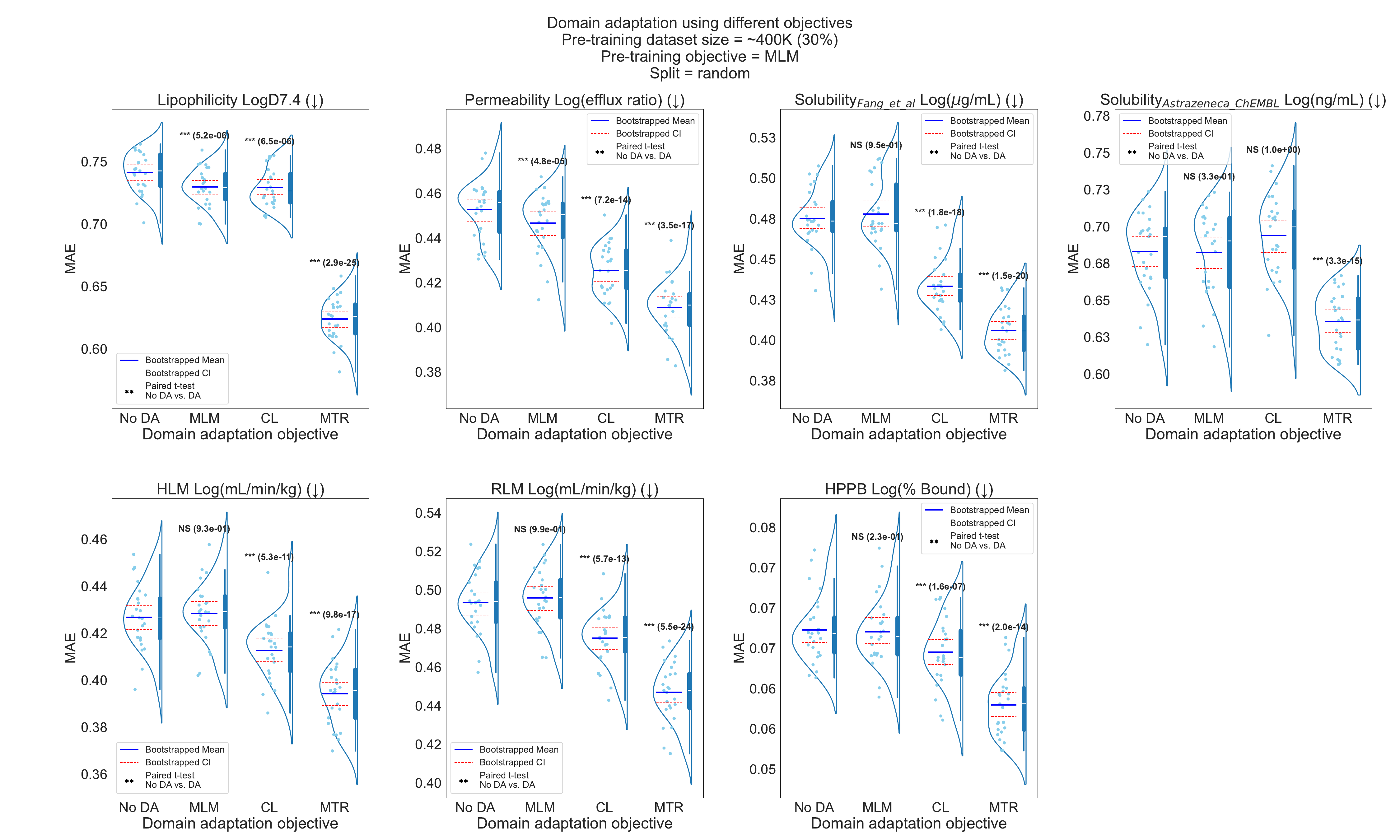}
    \caption{MAE performance for a baseline model trained with pre-training only (No DA), and three models incorporating domain adaptation (DA) using different objectives: Masked Language Modeling (MLM), Contrastive Learning (CL), and Multi-task Regression (MTR) for physicochemical properties. P-values are from one-tailed paired t-tests comparing each DA model to the No DA baseline, under the hypothesis that DA improves performance. Significance levels: * $p < 0.05$, ** $p < 0.01$, *** $p < 0.001$. Data is randomly split.}
    \label{fig:da_mae_random}
\end{figure}

\begin{figure}
    \centering
    \includegraphics[width=0.99\linewidth]{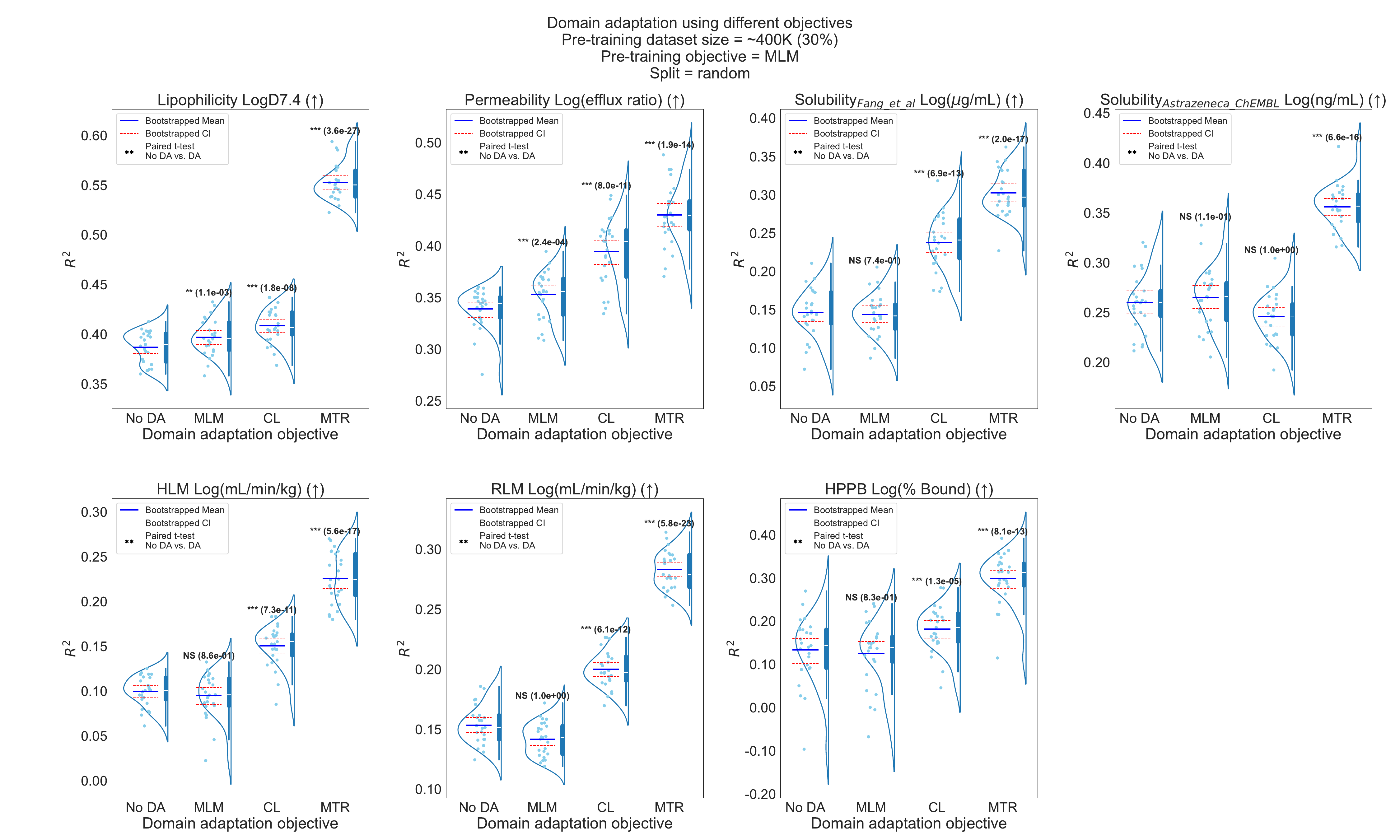}
    \caption{$R^2$ performance for a baseline model trained with pre-training only (No DA), and three models incorporating domain adaptation (DA) using different objectives: Masked Language Modeling (MLM), Contrastive Learning (CL), and Multi-task Regression (MTR) for physicochemical properties. P-values are from one-tailed paired t-tests comparing each DA model to the No DA baseline, under the hypothesis that DA improves performance. Significance levels: * $p < 0.05$, ** $p < 0.01$, *** $p < 0.001$. Data is randomly split.}
    \label{fig:da_r2_random}
\end{figure}

\begin{figure}
    \centering
    \includegraphics[width=0.99\linewidth]{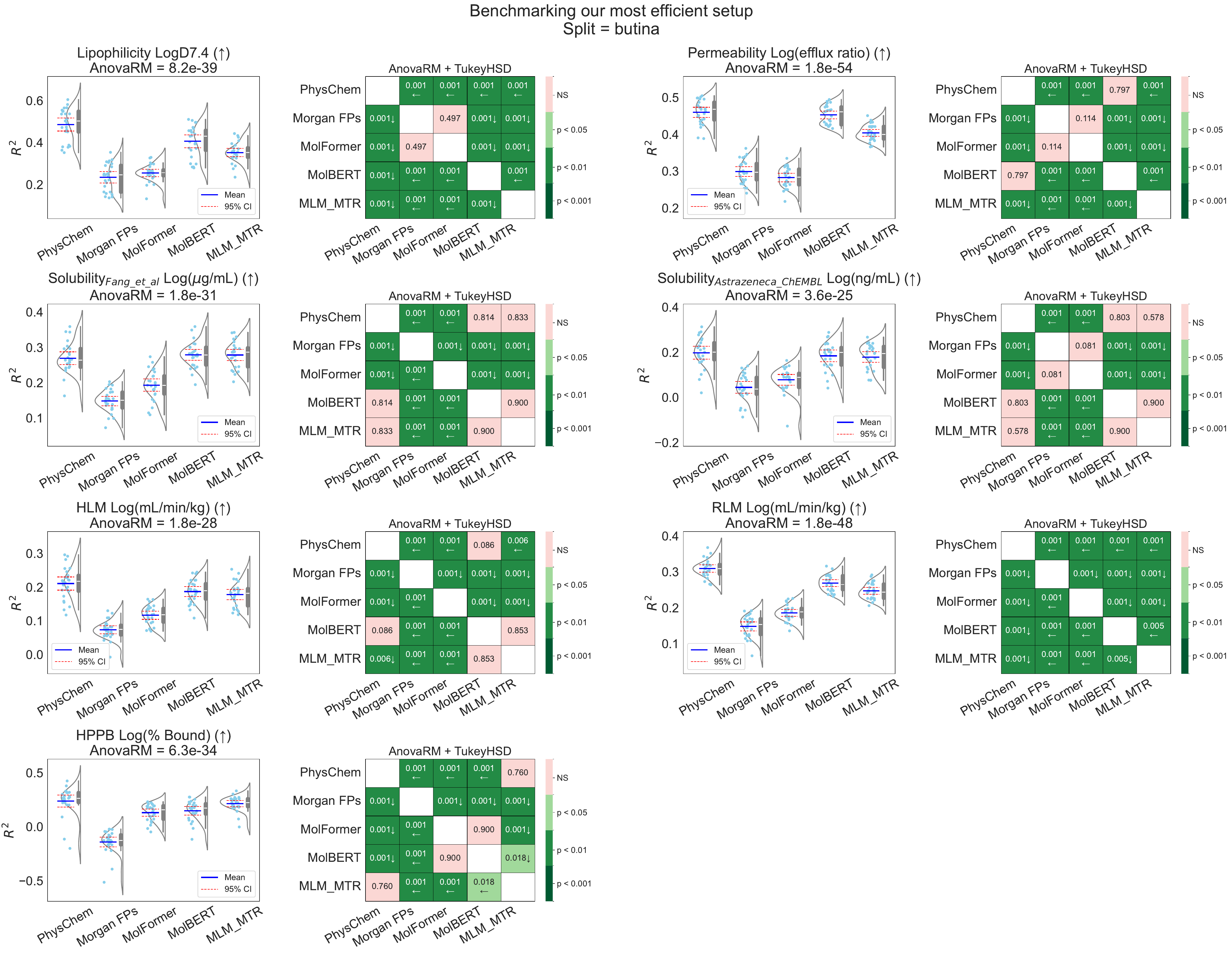}
    \caption{$R^2$ performance of the most efficient model from our analysis to models from the literature. MLM\_MTR corresponds to a transformer model pre-trained with $\sim 400$K molecules using the MLM objective then domain adapted on the corresponding endpoint using the MTR objective. MolBERT \cite{fabian2020molecular} is a chemically aware transformer than has been pre-trained on $\sim 1.3$M molecules using MLM, MTR, and SMILES-EQ objectives. MolFormer \cite{ross2022large} is a large-scale transformer pre-trained on 100M molecules using MLM. PhysChem and Morgan fingerprint correspond to two baselines using Random Forest models. Two-tailed significance analysis were performed, therefore, the arrows in the heatmap helps recognizing the model with the improved performance. CI = confidence interval for the estimation of the mean.}
    \label{fig:benchmark_r2_butina}
\end{figure}

\end{document}